\documentclass{article}
\usepackage{wrapfig}

\usepackage{graphicx}
\usepackage[final]{corl_wcbm_2024} 

\title{Learning to Look Around: Enhancing Teleoperation and Learning with a Human-like Actuated Neck}

\author{
  Bipasha Sen, Michelle Wang, Nandini Thakur, Aditya Agarwal, Pulkit Agrawal\\
  MIT Computer Science and Artificial Intelligence Laboratory\\
  \texttt{\{bise, wangmj, nsthakur, adityaag, pulkitag\}}@csail.mit.edu \\
}

\begin{document}
\maketitle


\begin{figure*}[h!]
    \centering
    \includegraphics[width=1\linewidth]{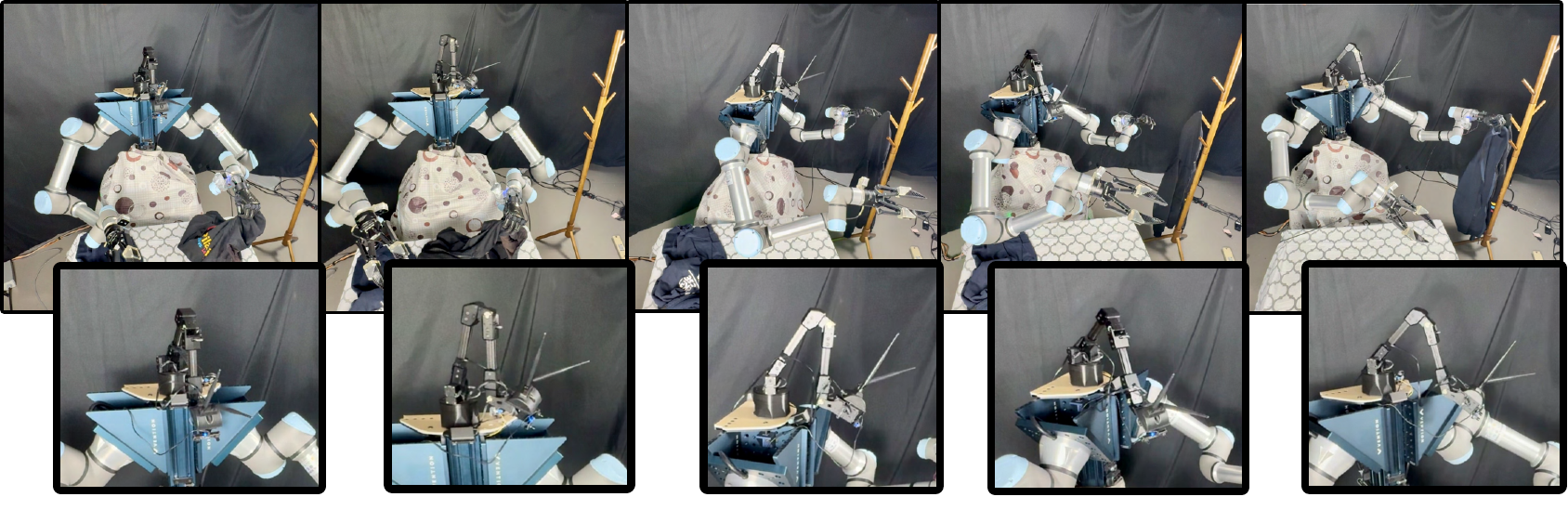}
    \vspace{-20px}
    \caption{\small A teleoperation system featuring an actuated neck and dexterous arms, enabling human-like manipulation in complex environments. The robot's 5-DOF neck mimics natural head movements, allowing behaviors like ``peeking" obstacles to locate objects. By incorporating human-like perception, the system broadens task capabilities and enhances efficiency, paving the way for advanced remote teleoperation.}

    \label{fig:teaser}
\end{figure*}

\begin{abstract}
We introduce a teleoperation system that integrates a 5-DOF actuated neck, designed to replicate natural human head movements and perception. By enabling behaviors like ``peeking" or ``tilting", the system provides operators with a more intuitive and comprehensive view of the environment, improving task performance, reducing cognitive load, and facilitating complex whole-body manipulation. We demonstrate the benefits of natural perception across seven challenging teleoperation tasks, showing how the actuated neck enhances the scope and efficiency of remote operation. Furthermore, we investigate its role in training autonomous policies through imitation learning. In three distinct tasks, the actuated neck supports better spatial awareness, reduces distribution shift, and enables adaptive task-specific adjustments compared to a static wide-angle camera.

\end{abstract}

\keywords{teleoperation, actuated neck, imitation learning} 


\section{Introduction}
Collecting high-quality demonstration data to train robotic manipulation algorithms presents significant challenges. Various methods for data collection have emerged, including the use of exoskeletons~\cite{exoskeleton, ace}, handheld manipulation devices~\cite{umi, umi_on_legs}, and retargeting human actions from videos into robot trajectories~\cite{imitation_body, imitation_hand, singh2024hand}. However, these approaches often involve translating the collected data into robot trajectories, which can introduce errors.

Alternatively, teleoperation~\cite{mobile_aloha, aloha, bunny_visionpro, aloha_unleashed, lin2024learning, chen2024arcap} offers a more direct data collection method, eliminating the need for retargeting. Yet, teleoperation poses its own set of challenges, particularly in terms of intuitiveness and ease of use for operators~\cite{teleop_issues}. Operators typically struggle with a limited field of view, especially if they are not positioned directly above the robot or constrained by a fixed camera setup such as in~\cite{open_television}, or lack haptic feedback. In this work, we address the challenge of perception gap and propose using a first-person perspective teleoperation by incorporating an actuated arm as a ``neck" that mimics natural human head movements, making teleoperation more intuitive.

\subsection{Human Perception and Intuitive Teleoperation}
In daily life, humans use head and neck movements extensively to navigate complex spaces and manipulate objects. A teleoperation system that replicates these natural movements can reduce cognitive load and enhance operator efficiency. To facilitate this, we mount a camera on an actuated neck with 5 degrees of freedom (DOF), allowing it to replicate human-like neck motions, including looking around obstacles and altering viewpoints through combined translation and rotation.

The increased DOF not only enables the operator to rotate and tilt the neck camera but also to "peek" around occlusions and create new viewpoints. This setup mimics the way humans perceive their environment, ultimately making teleoperation feel more natural and reduce the operator’s cognitive effort. As shown in Figure~\ref{fig:teaser}, adding the ability to peek around allows the teleoperator to perform complex whole-body maneuvers like finding the coat hanger relative to the viewpoint. This capability is especially crucial for whole-body remote teleoperation, where the operator is not physically present alongside the robot.

\subsection{Autonomy and Improved Data Collection}
In addition to improving the operator's experience, the actuated neck also has implications for autonomous policy training. Unlike static wide-angle cameras that often capture distorted images and can contribute to out-of-distribution perception, the actuated neck provides a dynamic viewpoint, reducing such risks. By using standard RGB cameras, image quality is enhanced, resulting in better data for both teleoperation and machine learning algorithms.

The dynamic camera adjustment also helps manage occlusions, allowing the robot to "peek" around obstacles to locate objects of interest, mirroring natural human behavior. In one task, for example, adjusting the camera’s angle was critical to finding a target object that was not visible in the direct line of sight. While a local wrist-mounted camera provided fine-grained details necessary for grasping or aligning the gripper, the actuated neck camera delivered crucial contextual information, aiding navigation within a broader environment.

The actuated neck also enables interactive perception~\cite{interactive_percep}, as it integrates sensor observations and action trajectories over time. This combined signal contains learnable and generalizable relationships, such as the ability to track objects as they move through space. By aligning the neck camera’s movement with object trajectories, we can implicitly learn which objects are of interest for specific tasks.

Our experiments trained a multi-task policy across different environment heights, demonstrating the neck’s ability to improve generalization and adaptability. The system effectively managed diverse scenarios, such as working on high and low tables, showing the potential for broader application.

Our key contributions are:

\begin{enumerate}
    \item Intuitive Teleoperation: We introduce an actuated neck with 5 degrees of freedom (DOF), allowing for human-like head movements. This design reduces operator cognitive load by enabling natural viewpoint adjustments.
    \item Improved Data Quality: The 5-DOF neck, equipped with standard RGB cameras, enhances image quality and minimizes perception errors, leading to better generalization for autonomous policy training.
    \item Enhanced Interactive Perception: The neck’s full range of motion supports dynamic object tracking and effective manipulation across diverse tasks, closely mimicking natural human behavior.
\end{enumerate}

\section{Hardware}

\begin{wrapfigure}[19]{R}{0.7\textwidth}
  \centering
  \vspace{-26px}
  \includegraphics[width=0.68\textwidth]{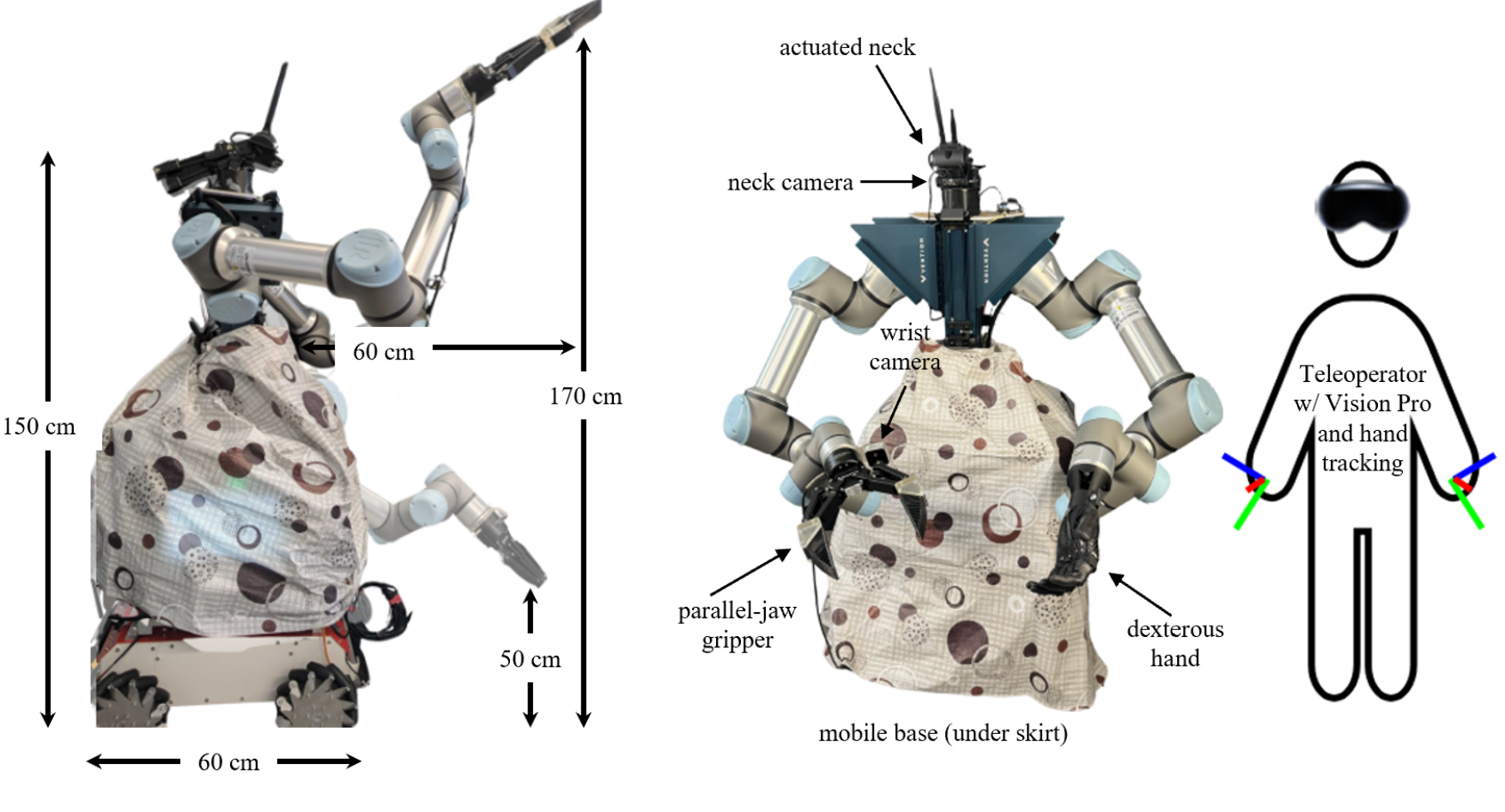}
  \caption{\small Overview of the teleoperated robotic system featuring a mobile base, 5-DOF actuated neck~\cite{interbotix}, dual UR5e arms~\cite{ur5}, and dexterous grippers~\cite{robotiq_gripper, psyonic}. The diagram details key dimensions, capabilities, and teleoperator setup, which includes Vision Pro for head tracking and hand tracking devices. The system is designed to provide human-like movement and precise control for remote teleoperation tasks, with a total of 21 degrees of freedom (14 for arms with parallel-jaw grippers, 5 for neck, and 2 for base) and additional 6 DoF with a Psyonic Ability dexterous hand.}
  \label{fig:hardware}
\end{wrapfigure}

To enhance teleoperation and ensure intuitive operation, our robotic system is designed to mimic human mechanics while balancing complexity, mobility, and robustness. We follow three core design principles: (1) \textbf{Human-like Mobility}: The system features an untethered mobile base, an actuated neck with 5 degrees of freedom, and dexterous arms, enabling human-like movements crucial for both local manipulation and global scene understanding. (2) \textbf{Remote Teleoperation Suitability}: The hardware is optimized for remote operation, enabling natural interactions even when operators are not physically present alongside the robot, crucial for intuitive control from a distance. (3) \textbf{Robustness}: The system’s arms are built to handle heavy payloads, and the base is designed for stability to prevent tipping during operation, ensuring safety and reliability.

The robot’s arms consist of two Universal Robotics UR5e arms~\cite{ur5}, chosen for their high payload capacity (5 kg each) and safety features like torque-based collision detection, allowing teleoperators to maneuver confidently without fear of causing damage. The arms have a reach ranging from 50 cm to 170 cm, enabling versatile manipulation across various heights.

For the neck, we use the Interbotix WidowX-200~\cite{interbotix}, which provides 5 degrees of freedom. This is a critical aspect of our design, allowing the neck to not only rotate and tilt but also to "peek" over and around obstacles through a combination of rotational and translational movements, mimicking human head dynamics as described above. This flexibility enhances operator perception, enabling better navigation of occlusions and complex environments during remote teleoperation.


The system’s torso is mounted on a metal frame that mimics a human’s upper body structure, as shown in Figure~\ref{fig:hardware}. The frame is secured to a Husarion Panther mobile base~\cite{husarion}, capable of speeds up to 2 m/s (comparable to brisk walking) and able to navigate uneven terrain. The base supports both differential and holonomic drive modes, providing additional flexibility for maneuvering in cluttered spaces. The wide base, combined with bottom-loaded weight from the mounted UR5 control boxes, prevents tipping, contributing to overall system stability.

Video streaming is enabled through four USB cameras positioned on the neck, torso, and wrists (using Intel Realsense D405 cameras~\cite{realsense}). The UR5 arms transmit proprioception data (joint positions, velocities, torques) over Ethernet, while the actuated neck communicates with the computer via serial connection. These sensory inputs are recorded and utilized to train imitation learning policies, aligning with our goal of integrating teleoperation data to support autonomous policy development.

\begin{figure}[t]
    \centering
\includegraphics[width=0.8\textwidth]{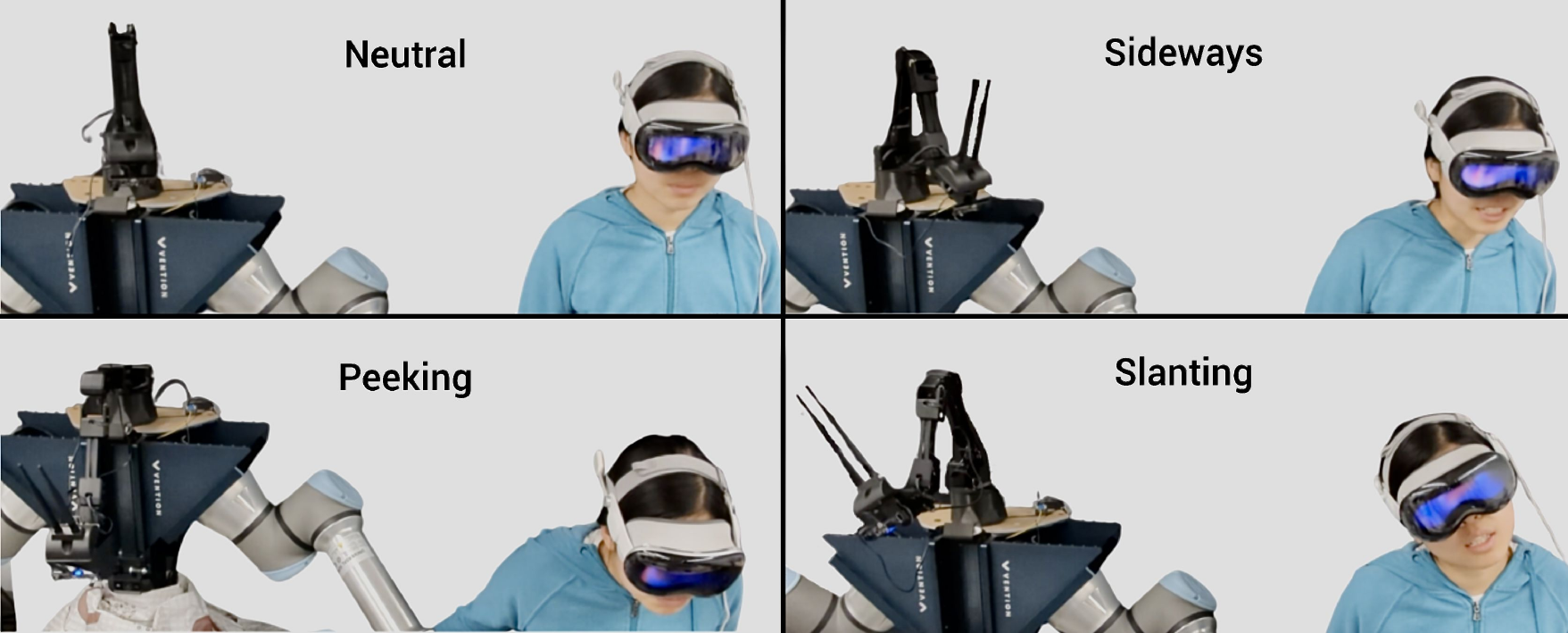}
    \caption{\small Demonstrating different head movements during teleoperation, mapped directly to the robot’s actuated neck: (Top left) Neutral, (Top right) Sideways, (Bottom left) Peeking, (Bottom right) Slanting. The robot’s 5-DOF neck enables natural, human-like adjustments, enhancing perception and control in complex environments.}
    \label{fig:teleop_active_neck}
\end{figure}

\section{Teleoperation and Data Collection}
Our teleoperation system consists of two main components: hand tracking and head tracking, both designed to provide a natural and intuitive interface for remote robot control. \footnote{The teleoperation code will be released.}

\subsection{Hand Tracking}
For hand tracking, we use the Ascension trakSTAR device—a 6DOF electromagnetic hand tracking system originally developed for teleoperating surgical robots~\cite{trakstar}. Tracking nodes are placed on the back of the teleoperator’s palm, thumb tip, and index finger tip using a glove. This setup is comfortable for the operator and enables precise capture of hand positions and finger movements, which are crucial for manipulating the robot's grippers.

To achieve fine-grained control of a dexterous robotic hand, we employ the Manus VR glove~\cite{manus}, which was initially developed for virtual reality gaming and provides 6DOF tracking of 25 keypoints on the hand. This allows detailed control of the robot’s dexterous psyonic hand~\cite{psyonic}, supporting complex manipulation tasks. Additionally, the gloves offer potential for future integration with haptic feedback systems, further enhancing the teleoperation experience.

We also conducted experiments using the Apple Vision Pro for hand tracking. While this approach provided promising results, it posed challenges for tasks performed close to the robot’s body, as the operator's hands sometimes moved out of the field of view of the headset. This led to tracking inconsistencies and jittering in the robot’s arms, which not only hindered task performance but could also be dangerous in close-proximity scenarios.

\subsection{Head Tracking}
6DOF head tracking is implemented using the Apple Vision Pro~\cite{park2024avp}. As shown in Fig~\ref{fig:teleop_active_neck}, changes in the operator’s head pose are directly mapped to the robot's neck camera pose, enabling a synchronized and intuitive perception experience. Video from the neck camera is streamed in real-time to the Vision Pro over a WiFi connection, allowing the operator to rapidly adjust the robot’s viewpoint, as humans naturally do when observing their surroundings. This supports interactive perception, enabling teleoperators to adapt to dynamic changes in the environment.

Unlike many existing teleoperation systems that rely on global third-person views or external feedback systems, our setup is fully self-contained. Visual feedback is conveyed exclusively through the Vision Pro, eliminating the need for global cameras or additional environmental feedback mechanisms. This streamlined design enables remote teleoperation in any location with a stable wireless connection, making it well-suited for diverse operational contexts.

\begin{figure}[t]
    \centering
    \includegraphics[width=0.9\textwidth]{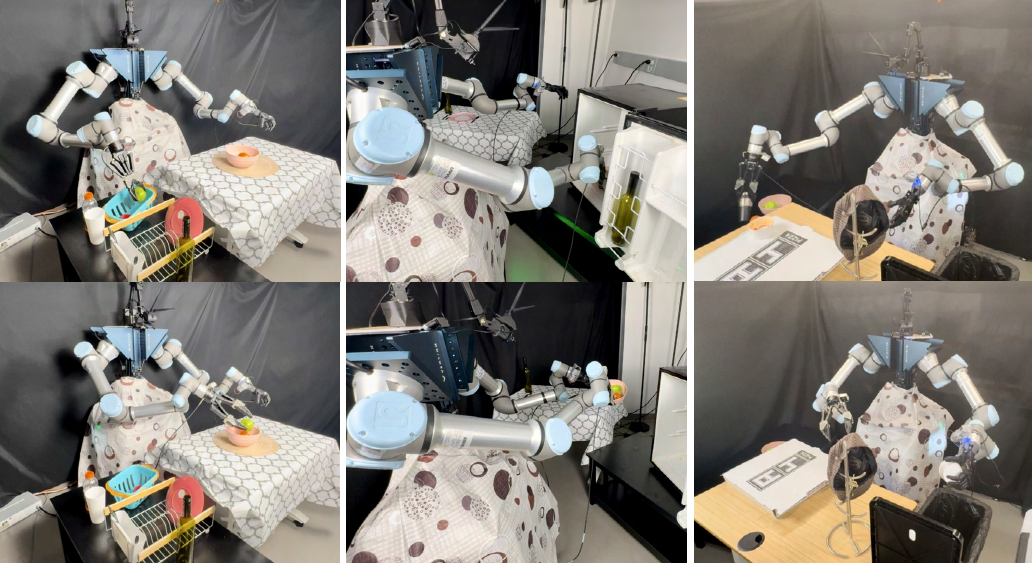}
    \caption{\small \textbf{Teleoperated Tasks: }The teleoperated robot performing various manipulation tasks, including setting up a dinner table (left), retrieving items from a refrigerator (middle), and managing workspace clutter (right). The 5-DOF actuated neck and dexterous hands enable precise control and adaptation to diverse environments, facilitating human-like interaction with household objects during teleoperation.}
    \label{fig:three_tasks}
\end{figure}

\section{Experiments}
We conduct a series of experiments to evaluate the effectiveness of the actuated neck in enabling intuitive remote teleoperation and to assess the capability of learning autonomous policies that leverage the actuated neck for task execution. 
We first highlight how the actuated neck tracks the teleoperator's head movements, as shown in Figure~\ref{fig:teleop_active_neck}, and enables a synchronized and intuitive perception experience for the teleoperator. Secondly, we train and evaluate the potential of training autonomous policies with the actuated neck, and compare it against a baseline policy trained using data from a static wide-angle camera without actuated neck movements. 

\subsection{Actuated Neck Enabled Intuitive Teleoperation}
To evaluate the impact of the actuated neck on teleoperation, we designed a series of seven complex, whole-body manipulation tasks that required coordinated use of a dexterous hand, a mobile base, and a parallel-jaw gripper. These tasks, (three of them visualized in Figure~\ref{fig:three_tasks}), included: (i) loading dishes into a dishwasher, (ii) hanging sweatshirts on a coat hanger, (iii) filling a cup with water from a faucet, (iv) arranging a dinner table, (v) making coffee, (vi) opening a refrigerator and transferring items from it to a table, and (vii) throwing trash into a bin. Each of these tasks demand dynamic perspective adjustments, precise dexterous manipulation, and base repositioning to achieve successful execution. Without an actuated neck, completing these tasks effectively would be extremely difficult, as they require the ability to change viewpoints, handle occlusions, and interact with objects from different angles.

The choice of tasks was deliberate, focusing on scenarios where the robot must use whole-body coordination, including base movement, to align with the workspace. For example, hanging a sweatshirt requires the teleoperator to not only locate the hanger but also move the mobile base to position the robot optimally relative to the hanger. This level of interaction makes dynamic viewpoint adjustments critical. The teleoperators, operating remotely, rely entirely on the robot’s onboard cameras, making the actuated neck indispensable for accurate perception. As shown in the coat hanger task in Figure~\ref{fig:teaser} and the fridge task in Figure~\ref{fig:three_tasks} (middle), the ability to "peek" around obstacles, observe objects from occluded angles, and adjust the head position enhanced situational awareness and task effectiveness significantly.

The actuated neck was especially important during dexterous manipulation, allowing teleoperators to refine the camera angle and achieve fine-grained observation of the dexterous hand’s movements. This led to improved control and task success, particularly in precision tasks such as filling a cup with water or placing items accurately on shelves. In contrast, using a wide-angle camera combined with software-based zooming would not have provided the same level of depth perception or ability to handle occlusions. Wide-angle lenses suffer from distortion and lack the ability to change perspectives, making it harder to adjust to dynamic environments or locate objects relative to the robot's body.

\begin{wrapfigure}[20]{R}{0.6\textwidth}
  \centering
  \vspace{-10px}
  \includegraphics[width=0.58\textwidth]{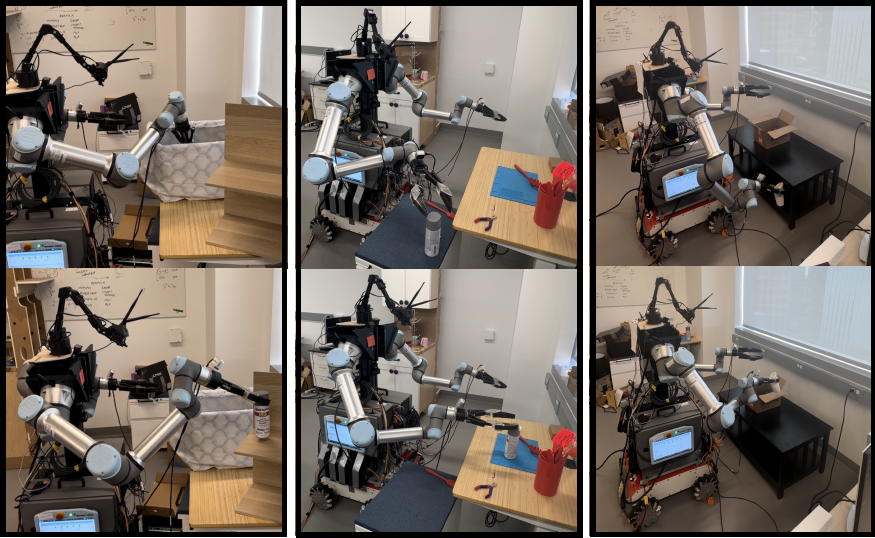}
  \caption{\small \textbf{Autonomous Tasks:}  A single policy is trained on three merged tasks across varying workspace heights and positions. (Left) The robot ``peeks" and picks up a cup, placing it on a top shelf. (Middle) The robot picks an object close to its body and places it on a table. (Right) The robot retrieves an item from under a coffee table and deposits it inside a package. The actuated neck adjusts dynamically based on observations, allowing the robot to identify and complete tasks without specific task labeling. }
  \label{fig:auttasks}
\end{wrapfigure}

\subsection{Impact of an Actuated Neck on Imitation Learning}

Building on the enhancements observed in teleoperation through the use of the actuated neck, we further investigate its potential to facilitate autonomous policy learning. Specifically, we aim to determine whether the actuated neck can enable policies that not only replicate the operator’s head movements but also reason about tasks at hand. We train an imitation learning policy, ACT~\cite{aloha}, that predicts joint configurations for the robot's right arm and the actuated neck. To train this policy, we concatenate the proprioceptive input, which includes joint positions of the actuated neck along with the arm and end-effector joints. Similarly, we combine the neck camera's observations with those from the wrist cameras to form the visual input. Since the evaluated tasks do not require long-term contextual reasoning, we set the context window to 1, focusing solely on immediate feedback. Additionally, to simplify the learning process, we concentrate on controlling the right arm with a parallel-jaw gripper while keeping the left arm fixed.

\begin{figure}[t]
    \centering
    \includegraphics[width=\textwidth]{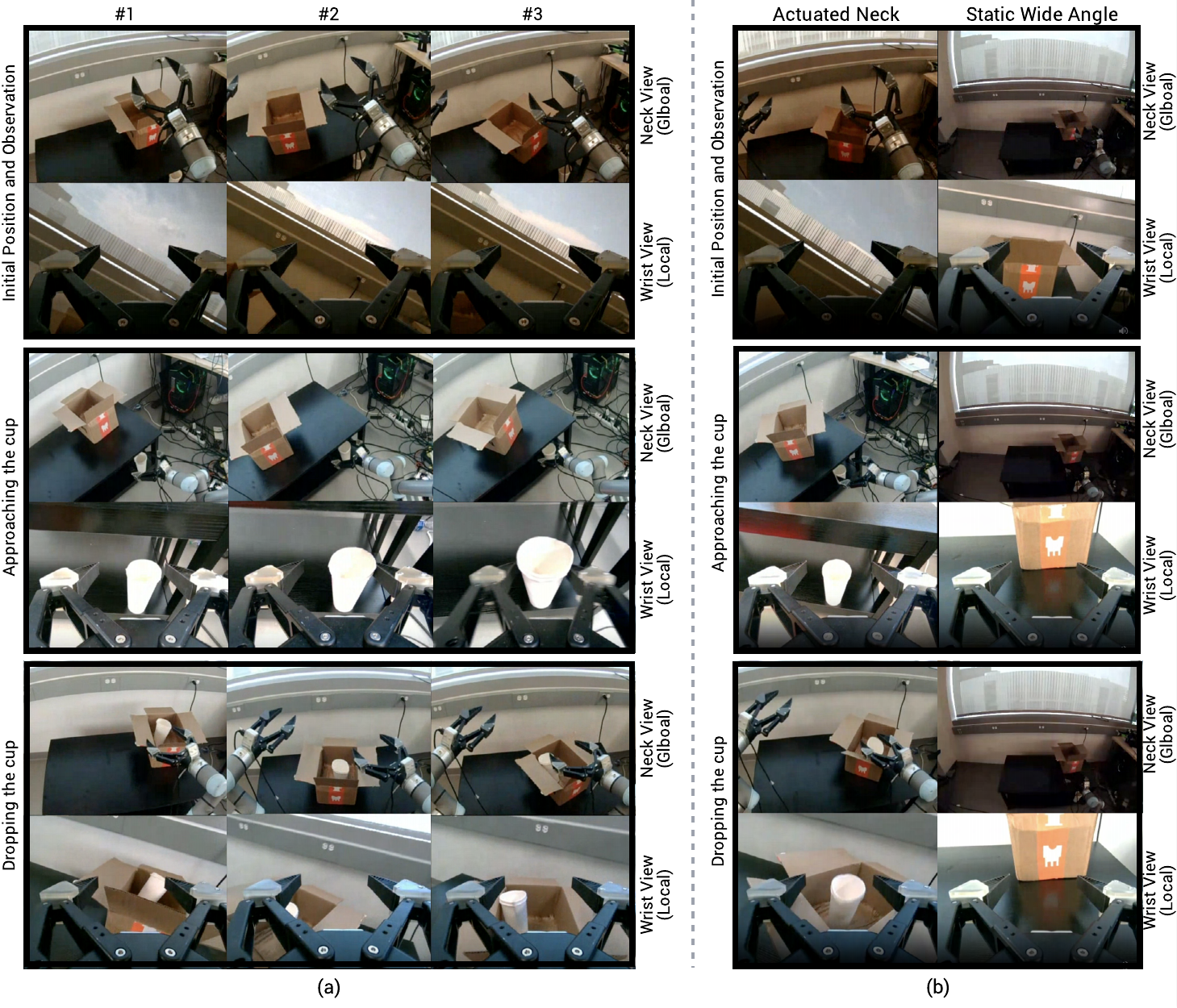}
    \vspace{-20px}
    \caption{\small \textbf{(a)} Visualization of the autonomous execution of the task \textbf{CfB} across three different instances (columns \#1, \#2, \#3). The actuated neck policy successfully generalizes to the three different instances of the task, despite the varying positions of the cup and the box highlighting its robustness.  
    \textbf{(b)} Compares the autonomous execution of the task CfB with the actuated neck policy and the static wide angle lens policy. In the left column, the policy with the actuated neck successfully completes the task, dynamically adjusting the neck to guide the robot's movements and place the cup inside the box. In the contrast, the robot with the wide angle policy fails to accomplish the task with the gripper getting stuck during the approach phase. }
    \label{fig:robotviewauttasks}
\end{figure}

To evaluate the role of the actuated neck in task reasoning and coordinated movement, we designed three specific tasks that required dynamic neck adjustments: (1) \textbf{Left-to-right Pick and Place (L2R):} The robot picks an object from a bin by peeking into it, then places it on a top shelf. This task requires precise horizontal neck adjustments for object localization, tracking, and guiding placement. (2) \textbf{Close-range Object Manipulation (CRange): }The object is initially located close to the robot's body, requiring the neck to retract inward to observe the object closely, enabling fine control for manipulation. (3) \textbf{Cup Transfer From Bottom Shelf (CfB):} The robot picks a cup from beneath a coffee table and places it in a packaging box on a higher table. This task necessitates vertical neck adjustments to maintain visibility of the cup throughout the transfer. 

\begin{wraptable}{r}{0.4\textwidth}
    \centering
    \begin{tabular}{|l|c|c|}
        \hline
        \textbf{Task $\uparrow$} & \textbf{Actuated} & \textbf{Static} \\
        \hline
        CfB & \textbf{95\%} & 0\% \\
        L2Rmod & \textbf{90\%} & 78\% \\
        CRange & \textbf{82\%} & 68\% \\
        \hline
    \end{tabular}
    \caption{\small Success rates of the actuated neck policy (Actuated) versus the static wide-angle camera policy (Static) across the three different tasks as described in Sec.~\ref{sec:wideangle}.}
    \label{tab:success_rates}
\end{wraptable}

These tasks, illustrated in Figure~\ref{fig:auttasks}, require dynamic spatial awareness and viewpoint adjustments, making them ideal for testing the active neck’s contribution to autonomous learning. We collected 120 teleoperated demonstrations for each task, simultaneously controlling the right arm and the neck. To assess whether the active neck goes beyond merely mimicking the operator’s movements—e.g., moving left-to-right without reasoning for the L2R task—we merged the datasets from all three tasks without explicit labeling and trained a single imitation learning policy. We let the policy figure out which task is at hand. To further verify the reasoning capability of the learned policy, we varied the location of the objects of interest during training and testing. Based on the learned policies, we make the following observations: 

\textbf{Global camera guides the local wrist cameras}: The wide workspace of these tasks necessitates reasoning over a broad visual area. The global camera, provided by the actuated neck, plays a critical role in guiding task execution. For example, in the CfB task, the neck adjusts to detect and localize the cup on the bottom shelf—an area initially outside the local camera’s view. Similarly, the neck aids the robot in correctly positioning the arm above the packaging box on the top table, as shown in Figure~\ref{fig:robotviewauttasks} (a), even when the box is not directly in view from the local cameras just after the robot successfully picks up the cup from the bottom of the coffee table. 

Conversely, the local wrist cameras handle finer adjustments, such as grasp alignment and release timing. For example, when approaching the cup in the CfB task, the wrist cameras refine the approach angle for a precise grip (see Figure~\ref{fig:robotviewauttasks} (a), "Approaching the cup"). Similarly, during the release phase, the wrist cameras ensure the gripper only lets go when the cup is correctly positioned inside the packaging box (see Figure~\ref{fig:robotviewauttasks} (a), "Dropping the cup").

\label{sec:wideangle}
\textbf{Comparison with a Wide-angle Camera:} A possible alternative to the actuated neck is a wide-angle camera, which could provide a broader field of view. To test this, we trained a separate policy using a 160-degree diagonal ultra-wide camera, following the same strategy of collecting 120 demonstrations across the three tasks. We modified the L2R task (L2Rmod) by removing the "peeking" behavior. Instead of picking the object from inside of a container, we simply place the object on the left side of the table to eliminate occlusions, focusing purely on horizontal movement. This is important as unlike the actuated neck that can adjust itself to observe objects not directly in line of sight, the static cameras cannot do that. We conducted 15 trials per task, slightly altering the object locations in each trial to assess the robustness of the learned policies. In each trial, we first test the actuated neck, and then bring the object back to the exact same position to test the static camera. 

As shown in Table~\ref{tab:success_rates}, the actuated neck significantly outperforms wide-angle cameras across tasks. This improvement is likely due to reduced distribution shift. While wide-angle cameras capture a broad visual field, this often causes a shift even if any minor detail change (such as the position of wires on the floor). In contrast, the actuated neck maintains a focused, dynamic view, adapting to specific areas and ensuring consistent visibility.
The performance gap is most notable in the CfB task, where the wide-angle camera fails completely. Here, the cup is located at the bottom of the frame, as shown in Fig.~\ref{fig:robotviewauttasks} (b). The static camera cannot adjust its view downward and  the cup is only visible at the bottom of the global image, causing the local cameras to get stuck at the table edge. Conversely, the actuated neck adjusts its height to keep the cup in view, guiding the local cameras to complete the task.
In the other two tasks, where objects are more centrally located, the wide-angle camera performs better but still lags behind the actuated neck. This emphasizes the importance of dynamic viewpoint adjustments, which enable the neck to maintain clear visibility and achieve higher success rates, particularly in occluded or boundary-positioned scenarios.

\section{Conclusion}
Collecting large amounts of high-quality demonstration data remains a challenging problem when building autonomous manipulation systems. Teleoperation is a promising approach for data collection, but can be challenging and non-intuitive for human operators. In this work, we introduce a teleoperation system that incorporates an actuated neck to mimic operator's head movements, enhancing both intuitive remote teleoperation and the training of autonomous policies. Our system reduces operator cognitive load, improves situational awareness, and enhances task efficiency, particularly in complex environments with occlusions. 

Through a series of experiments involving seven different tasks, we demonstrated that the actuated neck allows operators to perform whole-body remote teleoperation effectively, providing an intuitive and seamless control experience. 
Further, we investigate the role of the actuated neck in imitation-learning based autonomous policies by training a single policy across three different tasks. We observe that the actuated neck reasons about the tasks and adjusts its neck movements accordingly. Moreover, the active neck significantly enhances policy generalization and robustness, and significantly outperformed those trained with a fixed wide-angle camera. Summarily, the actuated neck not only enhances the operator's situational awareness but also contributes to more effective autonomous task execution by providing dynamic viewpoint adjustments to better facilitate perception and manipulation. 

In the future, we aim to extend the training of autonomous policies to a broader range of tasks enabled by the actuated neck.

\acknowledgments{We would like to thank Branden Romero for initial assistance in assembling the robot arms and Shankara Narayanan in helping collect robot demos. We would also like to thank the reviewers for their useful comments and feedback.}



\bibliography{example}  

\end{document}